\documentclass{article}
\usepackage{arxiv}
\usepackage[numbers, compress]{natbib}
\usepackage[utf8]{inputenc} % allow utf-8 input
\usepackage[T1]{fontenc}    % use 8-bit T1 fonts
\usepackage{hyperref}       % hyperlinks
\usepackage{url}            % simple URL typesetting
\usepackage{booktabs}       % professional-quality tables
\usepackage{amsfonts}       % blackboard math symbols
\usepackage{nicefrac}       % compact symbols for 1/2, etc.
\usepackage{microtype}      % microtypography
\usepackage[pdftex]{graphicx}
\usepackage{amssymb}
\usepackage{amsthm} % ADDED
\usepackage{caption}
\usepackage{amsmath, bbm}
\usepackage{subcaption}
\DeclareMathOperator{\ii}{\mathbf{i}}
\DeclareMathOperator{\jj}{\mathbf{j}}
\DeclareMathOperator{\kk}{\mathbf{k}}

\title{Rotation-Invariant Gait Identification with Quaternion Convolutional Neural Networks}

\author{
  Bowen Jing \\
  Stanford University \\
  \And
  Vinay Prabhu \\
  UnifyID\\
   \And
  Angela Gu \\
  Stanford University \\
  \And
  John Whaley \\
  UnifyID
}

\begin{document}
\maketitle
\begin{abstract}
A desireable property of accelerometric gait-based identification systems is robustness to new device orientations presented by users during testing but unseen during the training phase. However, traditional Convolutional neural networks (CNNs) used in these systems compensate poorly for such transformations. In this paper, we target this problem by introducing Quaternion CNN, a network architecture which is intrinsically layer-wise equivariant and globally invariant under 3D rotations of an array of input vectors. We show empirically that this network indeed significantly outperforms a traditional CNN in a multi-user rotation-invariant gait classification setting .Lastly, we demonstrate how the kernels learned by this QCNN can also be visualized as basis-independent but origin- and chirality-dependent trajectory fragments in euclidean space, thus yielding a novel mode of feature visualization and extraction.
\end{abstract}
\section{Introduction} 
\begin{figure}[h]
\centering
\includegraphics[width=1\textwidth]{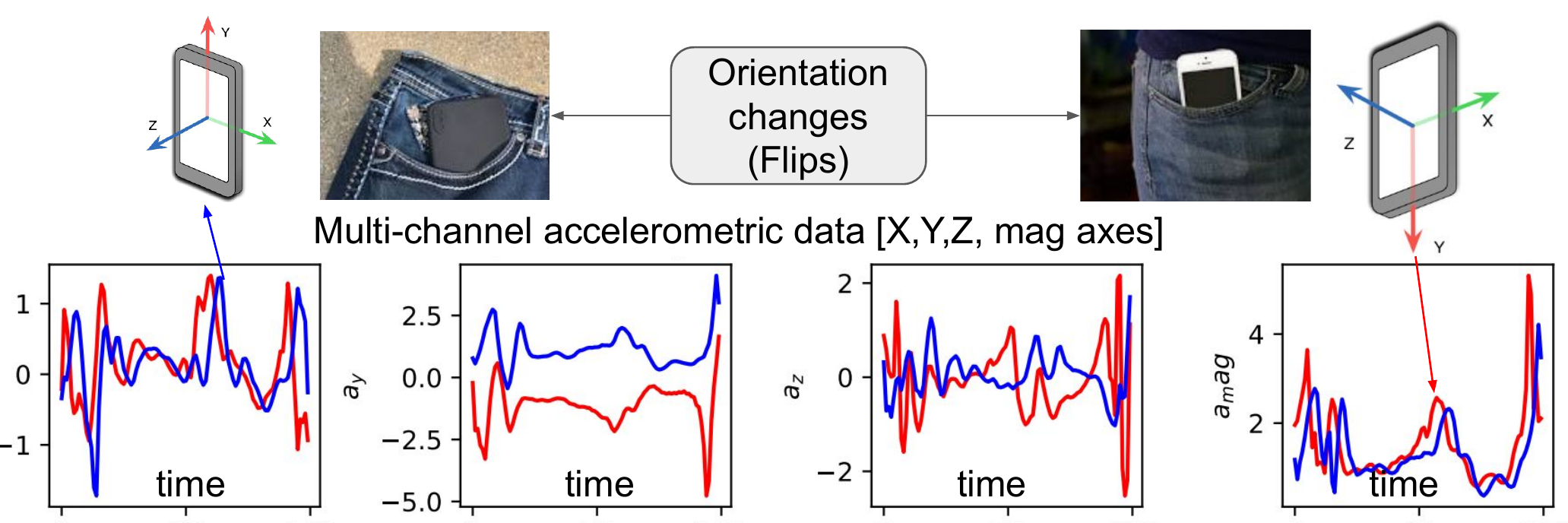}
\caption{Effects of an orientation flip on accelerometric measurements. Despite representing the same underlying gait and having very similar magnitude signatures (\textit{right}), the two gait cycles (red and blue) deviate significantly in their spatial $(x,y,z)$ components.}
\label{fig:intro}
\end{figure}

Accelerometric gait-based identification systems have increasingly embraced CNNs in lieu of hand-crafted features and shallow ML approaches \cite{gafurov2007survey_shallow}. These systems entail an \textit{enrollment} phase during which 3D accelerometric tensors of size $3 \times T$ are harvested after \textit{gait segmentation} \cite{idnet} and are used to train the CNNs. As long as the user maintains the same device orientation during the test phase, these models perform with high accuracy \cite{idnet}. However, they experience a catastrophic drop in accuracy if the user flips the orientation during the test phase. This device-flip and the resulting distributional shift (Figure \ref{fig:intro}) are typically tackled by either using only the acceleration \textit{magnitude}, or by explicitly performing rotation invariant transforms \cite{idnet}. The first option lowers the accuracy by discarding the rich 3D spatial information in the input tensors, and the second option is both computationally expensive (as it typically entails eigendecompostion of the accelerometric tensor in $\mathcal{O}(T^3)$) and adds to the software complexity of the system, especially for on-device implementations. A third option, data augmentation, often suffices in practice but substantially increases the complexity of the learning task presented to the network.

We instead incorporate rotation invariance \textit{inside} the CNN architecture by introducing a novel convolutional kernel which leverages quaternion representations of spatial rotations to learn $SO(3)$-equivariant maps between trajectory fragments in $\mathbb{R}^3$. Such kernels can then be stacked to form a network which is globally invariant to the orientation of input gait cycles. As desired, this network is agnostic to device orientations at both enrollment and test time, and instead learns to recognize basis-independent signatures of each user. These signatures, however, are significantly richer than those obtained by simply taking the magnitude of the accelerometric vectors.

Our work is in the same spirit as a number of recently developed $SO(3)$-equivariant architectures \citep{weiler20183d, thomas2018tensor, kondor2018n, zhang20193d, zhao2020quaternion} which have found promising applications in natural sciences as well as learning on 3D shapes. However, QCNN differs from many of these architectures in that it operates on arrays of vectors, not featurized point clouds or volumetric data---that is, it is \textit{not} permutation invariant, since input gait cycles are properly viewed as a vector time series, not a point cloud or shape. It also differs from a number of other quaternion neural networks \citep{zhu2018quaternion, parcollet2018quaternion} which leverage quaternion algebra to model non-linear transformations of real-valued, scalar data, but do not operate on spatial data or leverage quaternion algebra to encode spatial equivariances.

\section{Method} 

Our architecture which takes in a $n$-dimensional array of quaternions, performs a convolution-like operation on sliding windows, and outputs an array of quaternions. We use the trivial isomorphism between Euclidean vectors in $\mathbb{R}^3$ and pure quaternions
$$\begin{bmatrix}x \\ y \\ z\end{bmatrix}\in\mathbb{R}^3 \longleftrightarrow 0 + x\ii + y\jj + z\kk \in \mathbb{H}$$
to operate on input arrays whose elements are in $\mathbb{R}^3$. Similarly, a general quaternion can be thought of as a pair of a real number and a real vector in $\mathbb{R}^3$; we write this as $\hat{r} = r + \mathbf{r}$. When we say that the architecture is rotation-equivariant, we mean that it is equivariant to the rotation of the vector part of the quaternion.

Our architecture makes use of the representation of spatial rotations as the operation of \textit{quaternion conjugation}. Namely, if $\hat{r} = r + \mathbf{r}$, then $\hat{r}\hat{v}\hat{r}^{-1}$ yields a quaternion whose imaginary part corresponds to the vector $\mathbf{v}$ rotated by an angle $\theta = 2\arccos (r/|\hat{r}|)$ about the axis defined by $\mathbf{r}$. The real part of $\hat{v}$ is preserved under this operation.

\subsection{Convolutional kernel}

Consider a single-channel 1-D convolutional filter of length $2l+1$. The input to this filter is a window of quaternions $\hat{q}_0, \hat{q}_1, \ldots, \hat{q}_{2l}$. Call the input in the middle of the convolutional window $\hat{q}_l$ the \textit{pivot} of the convolution. Then the output of the filter for that window is
$$f(\hat{q}_0, \hat{q}_1, \ldots, \hat{q}_{2l}) = \sum_{i=0}^{2l} a_i(\hat{q}_i + b_i)(\hat{q}_l + c_i)\hat{q}_i(\hat{q}_i + c_i)^{-1}$$
where $a_i, b_i, c_i \in \mathbb{R}$ are the learnable parameters of the filter. This output is itself a quaternion, and is equivariant under 3D rotations of the \textit{vector part} of the input. That is, for any $\hat{r}\in \mathbb{H}$, 

$$\hat{r}f(\hat{q}_0, \hat{q}_1, \ldots, \hat{q}_{2l})\hat{r}^{-1} = f(\hat{r}\hat{q}_0\hat{r}^{-1}, \hat{r}\hat{q}_1\hat{r}^{-1}, \ldots, \hat{r}\hat{q}_{2l}\hat{r}^{-1})$$

This operation can be easily generalized to possess the same notions of stride, padding, multiple input channels, and multiple output channels as in standard real-valued convolutions. Assuming a filter length of $L$, $C_{in}$ input channels, and $C_{out}$ output channels, the kernel is characterized by $3LC_{in} C_{out}$ real-valued parameters.

Quaternion convolutional kernels can be stacked with no nonlinearity required between layers, as the composition of multiple convolutions is not reducible to a single convolution. These layers can then be joined to real-valued convolutional or dense layers to construct a deep, rotation-\textit{invariant} architecture by just taking the magnitude of the quaternion $\hat{q} \rightarrow |\hat{q}|$ or by taking its real part $\hat{q} \rightarrow (\hat{q} + \hat{q}^*)/2$.

\subsection{Normalization and initialization}
Although the kernels defined above fully specify the architecture of Quaternion CNN, some further considerations of normalization and parameter initialization are necessary for effective training. Consider a batch of $m$ 1-D inputs of length $n$ consisting of $C$ input channels. We define a \textit{normalized input} to mean a such batch whose RMS norm for each channel is one; that is $$\sqrt{\frac{\sum_{j=1}^m\sum_{i=1}^n|q^{(j)[k]}_{i}|^2}{mn}} = 1$$ for $k = 1, \ldots, C$. A \textit{quaternion batch norm} operation is defined as one which, with each forward pass during train time, updates an estimate of the RMS norm of the input for each channel $k = 1,\ldots, C$:
\begin{equation*}
    \mu^{[k]} := (1-\epsilon)\mu^{[k]} + \epsilon\sqrt{\frac{\sum_{j=1}^m\sum_{i=1}^n|q^{(j)[k]}_{i}|^2}{m n}}
\end{equation*}
with $\epsilon$ some small momentum term, and then outputs
$$q_i^{(j)[k]} := q_i^{(j)[k]}/\mu^{[k]}$$ for all $i, j, k$. This plays an analogous role to real-valued batch norm operations, although the implementation is different because a shift of the input would inject a non-rotation-equivariant translation into the data.

To initialize parameters, we consider the $a_i$, $b_i$, and $c_i$ separately. The weights $a_i$ play a role akin to the weights of a standard convolutional filter, and He initialization \cite{he2016deep} works well for producing near-normalized outputs. For the bias terms $b_i$, we make the simplifying assumption that the inputs in $\mathbb{H}$, when interpreted as vectors in $\mathbb{R}^4$, are drawn from a Gaussian distribution $\mathcal{N}(\mathbf{0}, I_4/4)$. Therefore we draw $b \sim \mathcal{N}(0, 1/4)$ to match the variance of the real part of the inputs. For rotations $c_i$, we want the resulting angles of rotation to be uniform $\theta \sim \text{U}[0, 2\pi]$. To achieve this, note that if the pivot quaternion $\hat{r} = r + \mathbf{r}$ has $|\mathbf{r}|^2=3/4$, as is the expectation, then the angle of rotation is $\theta = 2\arctan (3/(4r))$. Interestingly, the supplement of this angle as a function of $r$ is well approximated by the CDF of $\mathcal{N}(0, 1.3780^2)$. Therefore, if the real part of the rotation quaternion $q_l + c_i$ has variance $1.3780^2$, then the resulting angle distribution will be roughly uniform. Since the real part of the pivot quaternion already contributes variance $1/4$, we draw $c_i \sim \mathcal{N}(0, 1.3780^2 - 1/4)$.

\section{Experiments}

To investigate the performance of our method, we compare against standard CNNs on two learning tasks. In the cotemporal experiments, gait cycles are recorded from a small cohort of users\footnote{Here, and elsewhere, \textit{user} refers to research subject, not necessarily users of deployed UnifyID systems} carrying two simultaneous recording devices with opposite orientations. The data from one device is then used to train the networks, while the data from the other device is used for testing. This setup most closely mimics the real-world use case of users using one orientation to enroll, but presenting another orientation (of the same gait signature) to authenticate.

Due to the small size of the cotemporal dataset, however, we also perform multi-user experiments with a larger cohort of users for whom cotemporal data is not available. We mimick orientation shifts on this dataset by directly applying Euclidean rotations to the gait cycle tensors. We also apply such shifts to the training dataset in order to evaluate the efficacy of data augmentation with CNNs as compared with the intrinsically invariant QCNN.

In both experiments, raw accelerometric data is first segmented into gait cycles and each cycle is then resampled to be $T=100$ data points.

\subsection{Cotemporal experiments}
\begin{figure}[ht!]
    \centering
    \includegraphics[width=1\textwidth]{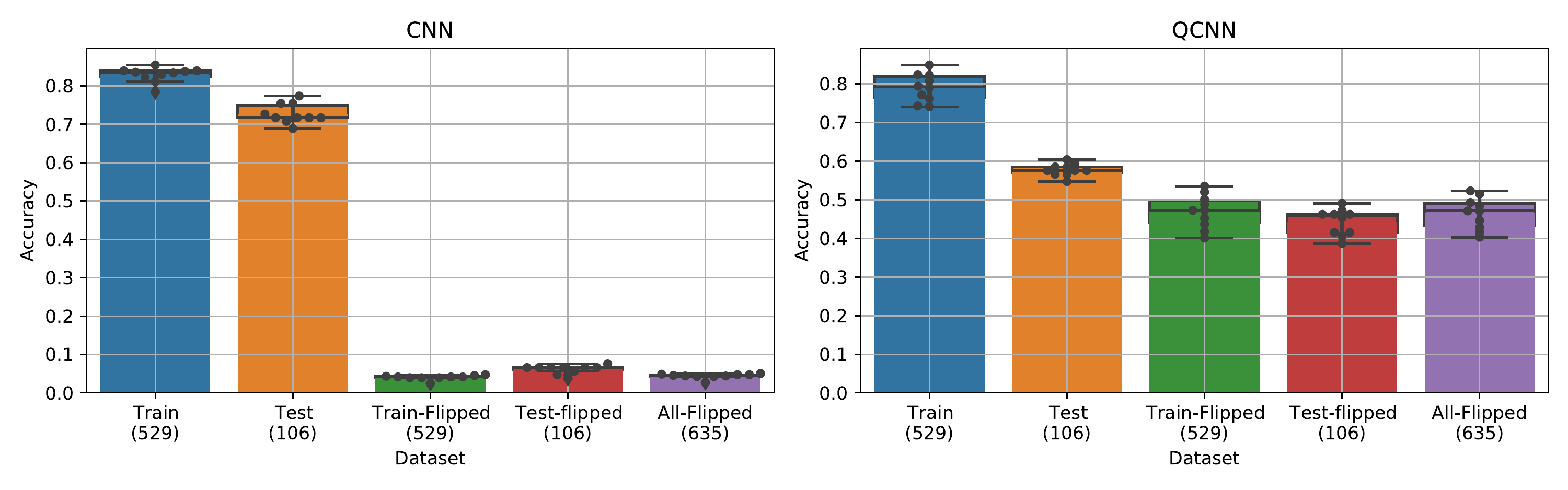}
      \caption{Classification accuracy of standard and quaternion CNNs over 10 trials of enrollment/testing. The CNN suffers a catastrophic drop in accuracy upon encountering unseen orientations, but the QCNN is more robust.}
\label{fig:Ccomparison}
\end{figure}
We enroll a cohort of eight volunteers (in a single orientation) who carried 2 phones (flipped/normally oriented) and collectively generated 635 gait cycles per phone. These cycles were then split 1:6 (529-train:106-test), where the normally-oriented "test" cycles were used for validation, and both splits of the "flipped" set were used for testing, but separately designated as "train-flip" and "val-flip." For a valid comparison, we use a 4-layer network for both the CNN and QCNN; the CNN has 213k parameters while the QCNN has 59k parameters.

The aggregate results from 10 independent train/test runs for each network are shown in Figure \ref{fig:Ccomparison}. As seen, when the standard CNN is tested on flipped data ("all-flipped"), there is a sharp drop in accuracy (78\% to 5\%) compared to the normally-oriented test data ("test"). No overfitting is observed on the "train-flipped" subset, confirming that the CNN is unable to translate even seen gait signatures from one orientation to another. The QCNN, on the other hand, has a 10x fold improvement over the CNN on the "all-flipped" test set (51\%). Additionally, as expected, the performance on "train-flipped" is sightly better than on "test-flipped," indicating that the overfitting to the training cycles has transferred to their flipped counterparts as well---an expected consequence of rotation invariance. However, looking \textit{only} at the normally-oriented data, the QCNN exhibits more overfit and thus lower performance on "test", suggesting that in the \textit{absence} of rotation invariance in the problem statement,\footnote{Strictly speaking, in the absence of rotation invariance \textit{and} in the absence of meaningful rotational information that can be used to distinguish the classes. These two criteria characterize problems where the inputs are all in some canonical orientation, as is the case here (the upright phone being the canonical orientation).} the QCNN may be less expressive than the standard CNN.

\subsection{Multi-user experiments}
In multi-user experiments, we use data collected from a cohort of 100 users consisting of 1000 training/validation cycles and 100 test cycles per user. Despite the large number of cycles per user, this dataset contains a limited number of rotation modalities per class.\footnote{These modalities nevertheless may change and a deployed system should not rely on them to identify users.} Therefore, to properly assess rotation-invariant identification we rotate each gait cycle by an random 3-D rotation to generate orientation-agnostic training and test sets. We then compare our model's performance to a standard CNN baseline when trained and tested on the rotated gaits (corresponding to training with data augmentation), when trained on the original gaits and tested on the rotated gaits (similar to the cotemporal setup), and when trained and tested on the original gaits (corresponding to the unlikely scenario of no new orientations at test-time).

As before, we compare standard and quaternion CNNs with the same number of layers (4 convolutional layers and 2 dense layers). However, since the number of classes is signicantly larger than in the cotempoeral experiments, most of the model parameters in our architecture are in the dense weights, so the reduction in parameter count in the QCNN is minimal (965892 reduced to 961300). In training both networks, we save the model with the highest top-1 validation accuracy (on the same type of dataset as the test set).

The test classification accuracies from the experiments are shown in Table \ref{tab:equivariance}. When the training and test sets are both in their original orientations, the standard CNN has higher accuracy, but again its performance drops precipitously when the test set is freely rotated. This trend appears similar to the trend in the cotemporal experiments, but its implication is subtly different: because the original orientations are not standardized and may differ from class to class, the CNN may be using rotation \textit{information} to distinguish different classes. Such information is destroyed when the test set is rotated, but is always unavailable to the QCNN. Importantly, data augmentation (rotated/rotated), which also destroys such information and forces the network to rely on rotation-invariant features, is unable to rescue the standard CNN to the performance level of the QCNN. Therefore, in the presence of rotation invariance, the CNN is less expressive than the QCNN, even when such invariance is injected into the training data. The latter's performance is essentially constant across the three datasets, a direct consequence of its \textit{intrinsic} rotation invariance.
\begin{table}[ht!]
    \centering
    \begin{tabular}{c|cc|cc}
    \hline
         &  \multicolumn{2}{c|}{QCNN} & \multicolumn{2}{c}{Standard CNN}\\
         Train/Test & Top-1 & Top-5 & Top-1 & Top-5 \\
         \hline
         Original/Original & 22.94\% & 33.28\% & \textbf{27.35\%} & \textbf{36.82\%} \\
         Original/Rotated & \textbf{22.94\%} & \textbf{33.28\%} & 8.36\% & 17.40\% \\
         Rotated/Rotated & \textbf{23.39\%} & \textbf{33.41\%} & 19.13\% & 29.32\% \\
    \hline
    \end{tabular}
    \caption{Test classification accuracies on the three types of datasets. As before, the standard network outperforms the quaternion network when all cycles are in the original orientation, but suffers a significant drop otherwise.}
    \label{tab:equivariance}
\end{table}

\subsection{Kernel visualization}

\begin{figure}[h]
  \centering
  \includegraphics[width=0.92\textwidth]{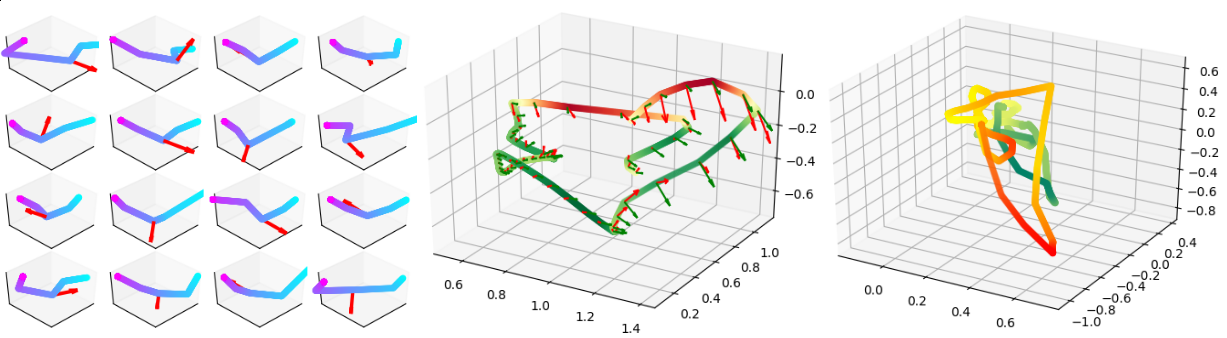}
  \caption{\textbf{Left:} Trajectory fragments which maximally activate the 16 kernels. The origin is shown at the intersection of the three planes. The red arrow is rooted at the pivot of the input and displays the vector part of the output. All trajectories shown are optimized from the same seed to give comparable orientations; other trajectories are possible for each kernel. \textbf{Middle:} The action of two kernels on an example gait cycle. The two filters are activated (as measured by the magnitude of the output) by different segments of the cycle, highlighted in red and green. The vector parts of the filter outputs are shown as red and green arrows. \textbf{Right:} The output of the two filters can also be viewed as two cycles in $\mathbb{R}^3$, with an additional dimension in the real part of the quaternion, indicated by the shade of color (stronger red or green corresponding to larger real part).}
\label{fig:kernels}
\end{figure}

The 3D nature of the accelerometric vectors which make up the input gait cycles enable the visualization of quaternion filters as \textit{trajectory fragments} in 3D space. In Figure \ref{fig:kernels} we visualize the features learned by each of the 16 kernels in the first quaternion layer of the network used in multi-user experiments. These features are defined with respect to the \textit{origin} and a \textit{chirality}, but independent of the axes. That is, they are rotation equivariant but \textit{not} translation or reflection equivariant. This is fundamentally different from the features detected by shape and point-cloud networks, which oftentimes posess all three spatial equivariances. While several kernels appear to correspond to similar input features, they may map them to different output features. Additionally, each kernel may recognize multiple features, giving output quaternions of similar magnitude but in different directions. Therefore, QCNNs can learn rich representations on the input space of gait cycles.

As a more concrete example, the application of two first-level kernels to an example gait cycle is shown is also shown in Figure \ref{fig:kernels}. The specific kernels shown correspond to numbers 8 and 11 in Figure \ref{fig:kernels}. The visualization highlights another difference with real-valued convolutions: although it is tempting to characterize segments of the input solely by the strength of kernel activations, the relative directions of activation also convey important information for the next layer.

\section{Conclusions}

We have presented an $SO(3)$-equivariant quaternion convolutional kernel and constructed a neural network specifically tailored for accelerometric gait classification invariant under a change of device orientation. This network outperforms standard convolutional networks of comparable depth on rotation-invariant gait identification, is parameter-efficient, and learns features which are easily visualized in 3D.

We anticipate that future work may focus on theoretical and empirical analyses of training, particularly addressing the issue of training stability, which we observed to be an issue in some experiments. Furthermore, developing faster implementation kernels would permit more effective architecture iteration. Finally, although for gait classification we have largely focused on 1D input arrays of vectors, the architecture easily extends itself to multidimensional vector arrays, and we anticipate that additional application areas for our architecture may emerge in the future.

\subsection{Ethics statement}
In this paper, we have described a technique that seeks to improve the real world efficacy of a privacy enhancing passive biometric system. While the promise of a \textit{friction-less experience} seems tempting, one needs to pay heed to the threat of the surveillance potential associated with this technique as well. Hence, in this regard, we would like to reemphasize that only those users who have been clearly educated about this technology and who have provided an active consent of data usage, be enrolled as part of the classification cohort. This has to be implemented both, at the User Interface (UI) level on the mobile device being used to mine the data as well as in the legal realm to ensure that all the requirements of the jurisdictional legislation (Ex: GDPR \cite{GDPR}) are being met.

\section*{Code}
The implementation of quaternion CNN is provided at \url{https://github.com/bjing2016/qcnn-pytorch}.

\section*{Funding}
BJ and AG are each supported by a UnifyID AI Fellowship.

\bibliographystyle{abbrvnat}

\small
\bibliography{references}
\end{document}